\documentclass{article}
\usepackage{spconf,amsmath,graphicx,multirow,hyperref,url}

\title{H-VECTORS: Utterance-level Speaker Embedding Using A Hierarchical Attention Model}
\name{Yanpei Shi\thanks{The first and second author contribute equally to this paper}, Qiang Huang, Thomas Hain}
\address{ Speech and Hearing Research Group\\
          Department of Computer Science, University of Sheffield\\
\texttt{\{YShi30, qiang.huang, t.hain\}@sheffield.ac.uk}}
\begin{document}
%
\maketitle
\begin{abstract}
In this paper, a hierarchical attention network to generate utterance-level embeddings (H-vectors)
for speaker identification is proposed. 
Since different parts of an utterance may have different contributions to speaker identities, 
the use of hierarchical structure aims to learn speaker related information
locally and globally. In the proposed approach, frame-level encoder and attention are 
applied on segments of an input utterance and generate individual segment vectors. 
Then, segment level attention is applied on the segment vectors to construct an utterance representation. 
To evaluate the effectiveness of the proposed approach,
NIST SRE 2008 Part1 dataset is used for training, and two datasets, Switchboard Cellular part1 and CallHome American English Speech, 
are used to evaluate the quality of extracted utterance embeddings on speaker identification and verification tasks. 
In comparison with two baselines, X-vector, X-vector+Attention, the obtained results show that 
H-vectors can achieve a significantly better performance. Furthermore, the extracted utterance-level
embeddings are more discriminative than the two
baselines when mapped into a 2D space using t-SNE.


\end{abstract}
\section{Introduction}\label{introduction}
The generation of compact representation used to distinguish speakers has been an attractive topic and widely used in 
some related studies, such as speaker identification \cite{park2018training}, verification \cite{snyder2017deep,novoselov2018deep,le2018robust}, 
detection \cite{mclaren2018train}, 
segmentation \cite{garcia2017speaker,wang2018speaker}, and speaker dependent speech enhancement \cite{chuang2019speaker,gao2015unified}.

To extract a general representation, Najim et al. \cite{dehak2010front} defined a ``total variability space''
containing the speaker and channel variabilities simultaneously, and then extracted the speaker factors by 
decomposing feature space into subspace corresponding to sound factors including speaker and channel effects.
With the rapid development of deep learning technologies, some architectures using deep neural networks (DNN) have been
developed for general speaker representation \cite{variani2014deep,snyder2018x}. 
In \cite{variani2014deep}, Variani et al. introduced the $d$-vector approach
using the LSTM and averaging over the activations of the last hidden layer for all frame-level features.
David et al. \cite{snyder2018x} used a five-layer DNN with taking into account a small temporal context
and statistics pooling. To further improve the performance for embedding generation, 
attention mechanisms have been also used in some recent studies \cite{wang2018attention, zhu2018self}.
Wang, et al. \cite{wang2018attention} used attentive X-vector where a self-attention layer was 
added before a statistic pooling layer to weight each frame. 

However, there might still need an improvement on how to highlight the importance of different part of the input utterance. 
For this issue, a hierarchical attention mechanism is employed in
this paper. This is inspired by Yang's work \cite{yang2016hierarchical} in document classification, 
where it claimed that not all parts of a document are equally relevant for answering a query
and attention models were thus applied to both word and sentence level
feature vectors via a hierarchical network.  
In the proposed approach, an utterance can be viewed
as a document, and its divided segments and acoustic frames are treated as sentences
and words, respectively.
An attention mechanism is then used hierarchically at both 
frame level and segment level. 
The utterance embedding can be constructed by first building
representations of segments from frames and then aggregating those into an utterance representation.
The use of this hierarchical attention network (HAN) can offer a way to obtain 
a discriminative utterance-level embedding by explicitly weight
target relevant features.

The rest of the paper is organized as follow: Section \ref{Theoretical Framework} 
presents the architecture of our approach. 
Section \ref{Experiment Setup} depicts the used data, experimental setup, and the baselines to be compared.
The obtained results are shown in Section \ref{Results}, and a conclusion is finally drawn in Section \ref{Conclusion}.

\section{Model Architecture}\label{Theoretical Framework}

\begin{figure}[th]
\centering
\includegraphics[height=8cm,width=7cm]{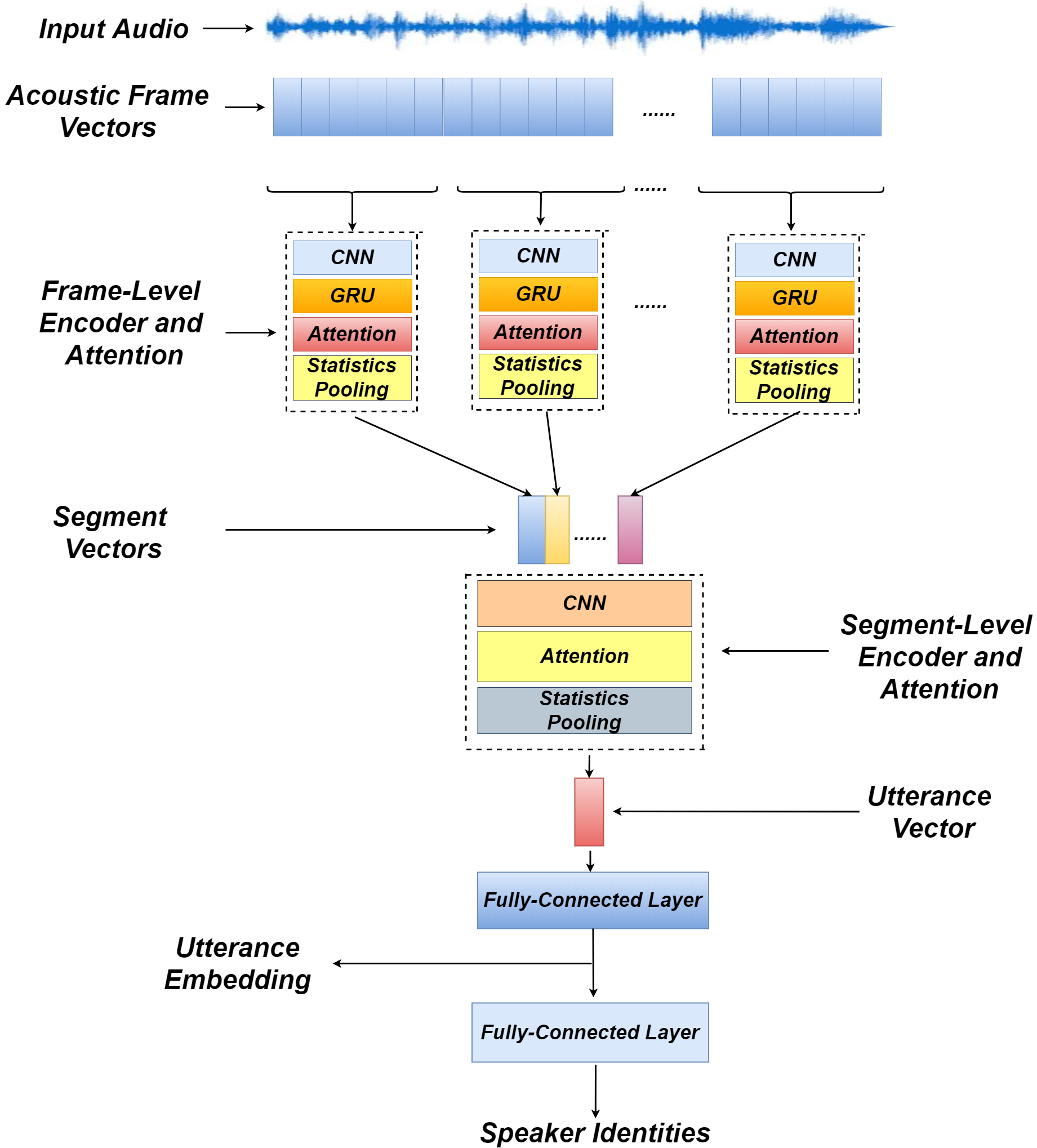}
\caption{Architecture of Hierarchical Attention Network.}
\label{proposed architecture}
\end{figure}

Figure 1 shows the architecture of hierarchical attention network.
The network consists of several parts: a frame-level encoder and attention layer,
a segment-level encoder and attention layer, and two fully connect layers.
Given input acoustic frame vectors, the proposed model generates
utterance-level representation, by which a classifier is trained
to perform speaker identification.  
The details of each part will be in the following subsections. 

\subsection{Frame-Level Encoder and Attention}
Assume that an utterance is divided into $N$ segments: $\textbf{S}=\{\textbf{S}_1, \textbf{S}_2, \cdots, \textbf{S}_N\}$
with a fixed-length window, and each segment $\textbf{S}_i = \{\textbf{x}_{i1}, \textbf{x}_{i,2}, \cdots, \textbf{x}_{i,M}\}$ constains $M$ $L$-dimensional acoustic frame vectors $\boldsymbol x_{i,t}$, 
$i \in \{1,\cdots N\}, t \in \{1, \dots, M\}$.

In the frame-level encoder, a one-dimensional CNN is used and followed by a 
bidirectional GRU \cite{chung2014empirical} in order to get information
from both directions of acoustic frames and contextual information.

\small{\centering{
$\boldsymbol x^{'}_{i,t} = \mathrm{CNN} (\boldsymbol x_{i,t})$

$\overrightarrow{h}_{i,t} = \overrightarrow{\mathrm{GRU}} (\boldsymbol x^{'}_{i,t})$

$\overleftarrow{h}_{i,t} = \overleftarrow{\mathrm{GRU}} (\boldsymbol x^{'}_{i,t})$

}}
\normalsize

The output of frame-level encoder $\boldsymbol h_{it}=[\overrightarrow{h}_{i,t}, \overleftarrow{h}_{i,t}], \in 
\mathcal {R}^{1 \times E}$
contains the summarized information of the segment centred around $\boldsymbol x_{i,t}$

In the frame-level attention layer, a two-layer MLP is first used
to convert $\boldsymbol h_{i,t}$ into a hidden representation $\boldsymbol z_{i,t}$, by which a
normalised importance weight $\alpha_{i,t}$ can be computed via a softmax function. 

\begin{equation}
\alpha_{i,t} = \frac{\mathrm{exp}(z_{i,t})}{\sum_{i=0}^{M} \mathrm{exp}(z_{i,t})}
\end{equation}

\begin{equation}\label{att}
z_{i,t} = \mathrm{Relu}(\boldsymbol h_{i,t} \boldsymbol W_{i,0}+\boldsymbol b_{i,0})\boldsymbol W_{i,1}
\end{equation}
where $\boldsymbol W_{i,0} \in \mathcal {R}^{E \times E}$, $\boldsymbol b_{i,0} \in \mathcal {R}^{1 \times E}$ and $\boldsymbol W_{i,1} \in \mathcal {R}^{E \times 1}$ are the parameters of the two-layer MLP.
These parameters are shared when processing $N$ segments.
A weighted sum of the output of frame-level encoder is computed by 
\begin{equation}
\boldsymbol A_{i,t} =  \alpha_{i,t} \boldsymbol h_{i,t}
\end{equation}
Following \cite{snyder2018x}, a statistics pooling is applied to $A_{i,t}$ to compute
its mean vector ($\boldsymbol \mu_{i}$) and std ($\boldsymbol \sigma_{i}$) 
vector over $t$.
A segment vector $\boldsymbol V_{S_{i}}$ is then obtained by concatenating the two vectors:
\begin{equation}
\boldsymbol V_{S_{i}} =  \mathrm{concatenate}(\boldsymbol \mu_{i}, \boldsymbol \sigma_{i})
\end{equation}

\subsection{Segment Level Encoder and Attention}

For segment-level attention, the same steps introduced in
frame-level encoder and attention are followed except a bi-directional GRU layer, 
as the omission of the GRU layer can well accelerate training when processing
a large number of samples.

%
%

\begin{table*}[h]
\renewcommand\arraystretch{1.0}
\centering
\begin{tabular}{cccccc}
    \hline
    Dataset & Type & \#Speaker & Size (hour) & \#Utterance (1s) &\#Utterance (3s)\\ \hline
    SRE08& Telephone+Interview & 1336 & 640 &3,528,326& 1,176,453\\ \hline
    CHE & Telephone & 120 & 60 & 252,224& 84,460\\ \hline
    SWBC & Telephone & 254 & 130 &1,008,901& 336,417 \\\hline

\end{tabular}
\caption{Details of three telephone speech datasets: Part1 of Sre2008 (SRE08), CallHome(CHE), and
Switchboard(SWBC).}  
\label{data}
\end{table*}

The weight $\alpha^s_n$ output of the segment-level attention layer can then be computed as follow \cite{pan2019automatic}:
\begin{equation}\label{att}
\begin{aligned}
z^{s}_{n} &= \mathrm{Relu}(\boldsymbol V_{n} \boldsymbol W_{n,0}+\boldsymbol b_{n,0})\boldsymbol W_{n,1}\\
\alpha^{s}_{n} &= \frac{\mathrm{exp}(z^{s}_{n})}{\sum_{n=0}^{N} \mathrm{exp}(z^{s}_{n})}
\end{aligned}
\end{equation}
where $\boldsymbol W_{n,0} \in \mathcal {R}^{E \times E}$, $\boldsymbol b_{n,0} \in \mathcal {R}^{1 \times E}$ and $\boldsymbol W_{n,1} \in \mathcal {R}^{E \times 1}$ are the parameters of a two-layer MLP used for .
A vector is generated using a statistics pooling over all weighted segments:
\begin{equation}
\begin{aligned}
   \boldsymbol \mu_{U} =  mean(\sum_n \alpha^s_n \boldsymbol S_n)\\
   \boldsymbol \sigma_{U} = std(\sum_n \alpha^s_n \boldsymbol S_n)\\
   \boldsymbol V_U = \mathrm{concatenate}(\boldsymbol \mu_{U}, \boldsymbol \sigma_{U})   
\end{aligned}
\end{equation}

The final speaker identity classifier is constructed using a two-layer MLP with $V_U$ being its input. 
As shown in Figure 1, the final utterance embedding $Emb_U$ is obtained after the first fully connected layer.

\section{Experiment}\label{Experiment Setup}
\subsection{Data}

Three datasets, NIST SRE 2008 part1 (SRE08), CallHome American English Speech (CHE), and Switchboard Cellular Part 1 (SWBC), are used in this paper to train
the proposed model and evaluate utterance embedding performance. 
SRE08 indicates the 2008 NIST speaker 
recognition evaluation test set \cite{sre08}, which 
contains multilingual telephone speech and English interview speech.
In this work, Part1 of SRE2008, containing about 640-hour speech and 1336 distinct
speakers, is selected in our experiments.    

SWBC \cite{swb} contains 130 hours telephone speech, totally 254 speakers 
(129 male and 125 female) under variance environment conditions (indoors, outdoors and moving vehicles). 
The stereo speech singles are split into two monos, and both of them are used in experiments. 
CHE \cite{callhome} contains 120 telephone conversations speech between native English speakers.
Among all of the calls, 90 of them are placed to various locations outside North America. 
In this dataset, speech from the left channel is used, as the labels of speakers 
in the right channels is unavailable. 
In our experiments,
SRE08 is used to train the proposed model, by which Utterance-level embeddings can be then generated using CHE and SWBC.

%
%
%

\subsection{Experiment Setup}

In this work, after removing unvoiced signals using energy based VAD \cite{pang2017spectrum}, fixed length sliding windows (one second or thre seconds) with half-size shift is employed 
to divide speech streams
into short segments. 
Each segment is viewed as an utterance independently. The total number of utterances
of the three datasets are listed in Table \ref{data}.
Each utterance is then split into 10 equal-length fragments without overlap.
Each fragment is further segmented into frames using a 25ms sliding window with a 10ms hop.  
All frames are converted into 20-dimensional MFCC feature vectors.
Similar to \cite{yang2016hierarchical}, to build a hierarchical structure,
each utterance, fragment and frame vector obtained here
are viewed as a document, sentence and word, respectively.

To evaluate the utterance-level embeddings, speaker identification and verification are conducted 
using the utterance-level embeddings generated on CHE and SWBC. 
Instead of directly processing on the embeddings, PLDA back-end \cite{Salmun2016OnTU} 
is applied on the embeddings to reduce the dimension to 300. 

Both SWBC and CHE datasets are randomly split into training and test
data with 9:1 ratio for speaker identification.
For speaker verification task, in SWBC, there are 50 speakers in the enrolment set and 120 speakers in the evaluation set,
with 10 utterances for each speaker.
In CHE, there are 30 speakers in the enrolment set and 60 speakers in the evaluation set. Each speaker has 10 utterances.

In order to compare the proposed approach with other speaker embedding systems, 
two baselines are built using the methods developed in previous studies. 
The first baseline ("X-Vectors") is based on a TDNN architecture \cite{snyder2018x}. It
is now widely used for speaker recognition and is effective in speaker embedding extraction.
The second baseline ("X-Vectors+Attention") is made by combining a global attention mechanism with a TDNN
 \cite{wang2018attention,zhu2018self}.
For evaluation, in our speaker identification task, correct prediction rate (prediction accuracy) is reported in this work. In the speaker verification task, equal error rate (EER) is reported. 
Moreover, to show the quality of the learned utterance-level embeddings, t-SNE \cite{maaten2008visualizing} is used
to visualize their distributions after being projected in a 2-dimensional space.

\begin{table}[h]
\renewcommand\arraystretch{1.0}
\centering  
\footnotesize
\begin{tabular}{c|c|c|c}
\hline
Level& Model & Input & Output  \\
\hline
\multirow{4}{*}{Frame-Level}&
CNN & (30,20,1) & (30,1,512)\\
&Bi-GRU&(30,512)& (30,1024)\\
&Attention & (30,1024)& (30,1024)\\
&Statistics Pooling & (30,1024)& (1,2048)\\
\hline

\multirow{3}{*}{Segment-Level}&
CNN & (10,2048,1) & (10,1,1500)\\
&Attention & (10,1500)& (10,1500)\\
&Statistics Pooling & (10,1500)& (1,3000)\\
\hline

\multirow{2}{*}{Utterance-Level}&
DNN & (1,3000) & (1,512)\\
&DNN& (1,512)& (1,512)\\
\hline

\end{tabular}
\caption{Architecture of the proposed approach}\label{model_summary}
\label{model_sum}
\end{table}

Table \ref{model_summary} shows the configuration of the proposed
architecture. It also contains batch normalisation \cite{ioffe2015batch} 
and droupout \cite{srivastava2014dropout} layers, where the dropout rate is set to 0.2.
Adam optimiser \cite{kingma2015adam} is used for all experiments with
$\beta_1=0.95$, $\beta_2=0.999$, and $\epsilon= 10^{-8}$. 
The initial learning rate is $10^{-4}$.


\section{Results}\label{Results}

\begin{figure*}[h]
\centering
\begin{minipage}[h]{.31\textwidth}
\centering
\includegraphics[width=4.6cm,height=3.2cm]{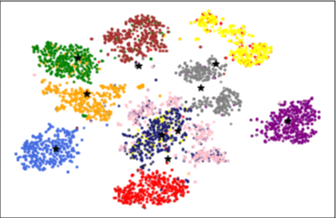}
\centerline{(a) X-vector}
\end{minipage}
\begin{minipage}[h]{.31\textwidth}
\centering
\includegraphics[width=4.6cm,height=3.2cm]{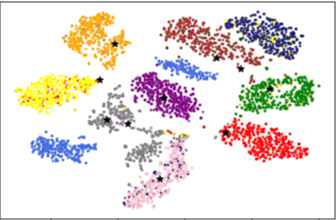}
\centerline{(b) X-vector+Attention}
\end{minipage}
\begin{minipage}[h]{.31\textwidth}
\centering
\includegraphics[width=4.6cm,height=3.2cm]{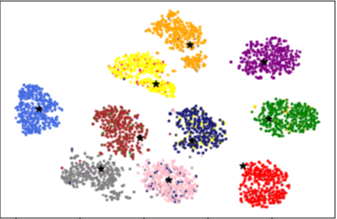}
\centerline{(c) H-vector}
\end{minipage}
\caption{Embedding visualization using t-SNE. Each color represents a speaker, 
and each point indicates an utterance. }
\label{tsne}
\end{figure*}



Table \ref{train3model} shows the prediction accuracies on the test data of SRE08 using
the proposed approach and two baselines.
Two different utterance lengths, 1 second and 3 seconds, are used in the experiments, respectively.
The use of the H-vectors shows higher accuracy when using either 1-second or 3-second input length
than the two baselines. 
When the length of input utterances is one second, the accuracy obtained using the H-vectors can reach 94.5\%, 
with 4.4\% improvement over X-vectors and 2.4\% improvement over X-vectors+Attention, respectively. 
When the length of input utterances is three seconds, the accuracy obtained using the H-vectors can reach 98.5\%, 
with about 3\% improvement over X-vectors and about 2\% improvement over X-vectors+Attention.
The proposed approach is more robust than the two baselines when processed utterances are short.
In addition, the accuracies obtained using 3-second utterances are better than those using 1-second utterances. 
As longer utterance contains more
information relevant to a target speaker than those in short ones.

\begin{table}[h]
\centering  
\footnotesize
\begin{tabular}{c|c|c}
\hline
Utterance Length& Model & Accuracy \% \\
\hline
\multirow{3}{*}{1 Second}&
X-vector & 90.1\\
&X-vector+Attention&92.1\\
&H-vector & 94.5\\
\hline

\multirow{3}{*}{3 Seconds}&
X-vector & 95.2 \\
& X-vector+attention &96.7\\
& H-vector & 98.5\\
\hline
\end{tabular}
\caption{Identification accuracy on the test data of SRE08 when the utterance length is 1s or 3s. Number of speakers is 1336}
\label{train3model}
\end{table}

\begin{table}[h]
\centering  
\footnotesize
\begin{tabular}{c|c|c|c}
\hline
Utterance Length& Model & Accuracy \% & EER \%\\
\hline
\multirow{3}{*}{1 Second}&
X-vector & 84.8 & 1.94\\
&X-vector+Attention&87.5& 1.61\\
&H-vector & 89.1& 1.44\\
\hline

\multirow{3}{*}{3 Seconds}&
X-vector & 89.4 & 1.46\\
& X-vector+attention &91.0& 1.21\\
& H-vector & 92.8&1.08\\
\hline
\end{tabular}
\caption{Identification accuracy and Equal Error Rate (EER) on CHE dataset when the utterance length is 1s or 3s.}
\label{callhome}
\end{table}

\begin{table}[h]
\centering  
\footnotesize
\begin{tabular}{c|c|c|c}
\hline
Utterance Length& Model & Accuracy \% &EER \%\\
\hline
\multirow{3}{*}{1 Second}&
X-vector & 78.2& 2.23\\
&X-vector+Attention&81.0& 2.05\\
&H-vector & 83.7&1.92\\
\hline

\multirow{3}{*}{3 Seconds}&
X-vector & 81.3 &2.01\\
& X-vector+attention &84.0&1.82\\
& H-vector &86.2&1.69\\
\hline
\end{tabular}
\caption{Identification accuracy and Equal Error Rate (EER) on SWBC dataset when the utterance length is 1s or 3s.}
\label{swb}
\end{table}

To evaluate the quality of embeddings extracted using the proposed approach, 
two additional datasets are employed in our experiments.  
Table \ref{callhome} and Table \ref{swb} show the identification accuracy and verification EER when using 
the embeddings extracted on SWBC and CHE dataset, respectively. 
On the two datasets, the H-vectors consistently outperforms the two baselines whether the length of utterances is one second
or three seconds. 

Since the model is trained on SRE08, the identification performances
on its test data are clearly better than those on the other two datasets.
As the SWBC dataset contains a wide range of environment conditions 
(indoors, outdoors and moving vehicles), both its identification and verification performances 
are relatively worse than those obtained on CHE dataset.

%

To further test the quality of extracted utterance-level embeddings, t-SNE \cite{maaten2008visualizing} is used to
visualise the distribution of embeddings by projecting these high-dimensional vectors into a 2D space.
From SWBC dataset, 10 speakers are selected and 500 three-second segment are randomly sampled for each speaker.
Figure 2 (a), (b), and (c) show the distribution of selected 3 seconds utterances of 10 speakers from SWBC dataset after using X-vectors, X-vectors+Attention, and H-vectors, respectively.
Each color represents a single distinct speaker and each point represents an utterance. 
The black mark represents the center point of each speaker class. 
Figure \ref{tsne}(a) shows the distribution of the embeddings obtained by X-vectors.
It is clear that, in this figure, some samples from different speakers are not well discriminated as there are overlaps between speaker classes.
Due to the use of an attention mechanism in X-vectors+Attention, 
figure \ref{tsne}(b) shows a better sample distribution than figure \ref{tsne}(a). However, some samples of a speaker
labelled by blue colour are not well clustered.
In figure \ref{tsne}(c), the embedding obtained by H-vectors performs better separation property
than the other two baselines.  




\vspace*{-2mm}
\section{Conclusion And Future Work}\label{Conclusion}
\vspace*{-2mm}
In this paper, a hierarchical attention network was proposed utterance-level embedding extraction. Inspired by the hierarchical structure of a document made by words and sentences, 
each utterance is viewed as a document, segments and frame vectors
are treated as sentences and words, respectively. 
The use of attention mechanisms at frame and segment levels provides a way to 
search for the information relevant to target locally and globally, and thus obtained better utterance level embeddings, including better performance on speaker identification and verification tasks using the extracted embeddings. Moreover, the obtained utterance-level embeddings are more discriminative than the use of X-vectors and X-vectors+Attention.


In the future work, different kinds of acoustic features such as filter-bank and Mel-spectrogram will be investigated and
tested on some large datasets, such as Voxceleb1 and 2.

\bibliographystyle{IEEEbib}
\bibliography{strings,refs}

\end{document}